# Physics-Aware End-to-End Deep Reinforcement Learning for Quadcopter Control with Actuator Dynamics

Ya-Chia Shen[a,b], Woei-Leong Chan[b]
[a] Department of Computer Science and Information Engineering, National Cheng Kung University
[b] Department of Aeronautics and Astronautics, National Cheng Kung University

## Abstract

Unmanned aerial vehicles (UAVs), particularly quadcopters, present unique challenges for autonomous control due to their underactuated dynamics—only four available control inputs must govern six degrees of freedom. This paper investigates a *physics-aware, end-to-end* deep reinforcement learning (DRL) approach that acts directly on low-level body inputs—total thrust and body torques $(T, \tau_x, \tau_y, \tau_z)$—and closes the loop through a high-fidelity Simulink environment. Our simulator integrates a 12-state rigid-body model (MATLAB Level-2 S-Function) with (i) an *Action2RPM* allocation based on the Moore–Penrose pseudo-inverse of a coefficient matrix derived from thrust and drag terms, and (ii) first-order actuator dynamics for each motor (time constant $T_m = 0.076s$) including rotor gyroscopic coupling. A shaped reward balances goal-reaching and stability using an exponential position well, attitude penalties, and quadratic velocity costs. Four DRL algorithms—DDPG, TD3, PPO, and SAC—are evaluated in two stages: (S1) thrust-only hover and (S2) hover with pitch torque and a translated goal. Results show that SAC and TD3 achieve superior stability and exploration efficiency, while PPO is less sample-efficient. The study highlights the significance of modeling actuator lags and aerodynamic moments for stable low-level control and provides a reproducible benchmark for quadcopter DRL.

## I. INTRODUCTION

Unmanned aerial vehicles (UAVs) present a challenging benchmark for continuous-control learning: they are underactuated, fast, and tightly coupled, with safety-critical dynamics and strict actuator limits. Deep reinforcement learning (DRL) has recently shown promise on both task-level navigation and low-level attitude/thrust control, leveraging function approximation to learn policies directly from state feedback [1]–[3]. For continuous control, prominent algorithms include DDPG [4], TD3 [5], PPO [6], and SAC [7], which trade off exploration, stability, and sample efficiency in different ways. While many studies validate in simplified simulators or body-level abstractions, bridging the sim-to-real gap requires models that respect thrust/torque generation and actuator dynamics [8]–[12].

**Problem and gap.** A large portion of UAV DRL research assumes idealized rigid-body inputs—e.g., instantaneous application of $\{T, \tau_x, \tau_y, \tau_z\}$ — omitting rotor aerodynamics, actuator lags, and allocation nonlinearities. Such assumptions can ease learning but obscure sim-to-real issues, particularly in yaw authority, cross-coupled responses, and closed-loop bandwidth limits. Recent progress in architectural priors (e.g., equivariance) [13], multi-agent coordination [14], and aggressive flight [12] does not fully expose policies to realistic actuation and mixing.

**This work.** We present a *physics-aware, end-to-end* DRL framework for quadcopter control, built in MATLAB/Simulink with a Level-2 S-Function plant. The agent outputs body-level actions $u = [T, \tau_x, \tau_y, \tau_z]^\top$, which are mapped to per-rotor commands through an explicit thrust/torque allocation and first-order motor dynamics, including gyroscopic coupling. This Simulink-in-the-loop design preserves control authority and bandwidth constraints encountered on hardware.

**Staged learning protocol.** To disentangle exploration from stabilization, we employ a two-stage curriculum: an initial thrust-only hover stage with randomized states, followed by a stage that activates an additional torque channel and nontrivial waypoints. This exposes algorithmic differences in efficiency and stability while progressively exercising allocation and actuator dynamics.

**Contributions.**

- **Physics-aware end-to-end pipeline:** A Simulink-in-the-loop DRL framework that coupling body-level actions with explicit rotor allocation and motor dynamics.
- **Action2RPM allocation layer:** An interpretable *Action2RPM* layer parameterized by $c_t$, $c_q$, and geometry, enabling reuse across airframes.
- **Controlled comparison of DRL algorithms:** A fair evaluation of SAC, TD3, DDPG, and PPO on the same physics-aware plant, highlighting exploration, stability, and sample-efficiency trade-offs [4]–[7], [15]–[17].
- **Practical training details:** A reproducible training setup (observation design, reward shaping, termination logic, parallelization caveats) tailored to actuator-constrained UAVs, complementing recent robust/realistic controllers [8]–[11].

**Organization.** Section II reviews the fundamentals of continuous-control reinforcement learning; Section III summarizes related work on UAV control and highlights the gaps motivating our approach. Section IV presents the system formulation and closed-loop model; Section V describes the learning methodology; Section VI outlines the experimental setup and evaluation metrics; Section VII presents quantitative and qualitative results; Section VIII concludes the paper and discusses future directions in sim-to-real transfer and multi-agent extensions.

## II. PRELIMINARIES

This section summarizes the reinforcement learning (RL) concepts and notations used throughout, emphasizing continuous-control actor-critic methods.

### A. Markov Decision Processes and Values

We model the environment as a discounted MDP $\mathcal{M} = (\mathcal{S}, \mathcal{A}, p, r, \gamma)$ with policy $(\pi_\theta(a \mid s))$ and return

$$G_t = \sum_{k=0}^{\infty} \gamma^k r(s_{t+k}, a_{t+k}) \tag{1}$$

The action-value is

$$Q^\pi(s,a) = E_\pi[\, G_t \mid s_t = s, a_t = a\,] \tag{2}$$

We use standard Bellman targets implicitly through the updates below.

### B. Policy Gradient and Actor-Critic

Maximizing $J(\pi_\theta) = E_{s\sim d_\pi, a\sim\pi_\theta}[Q^\pi(s,a)]$, the stochastic policy gradient [2] is

$$\nabla_\theta J = E[\nabla_\theta \log \pi_\theta(a \mid s)\, Q^\pi(s,a)] \tag{3}$$

Actor-critic uses a critic $Q_\phi$ and a one-step TD target $y$ to fit $Q_\phi$ via $\mathcal{L}_Q = E\left[\left(Q_\phi(s,a) - y\right)^2\right]$.

### C. Deep Deterministic Policy Gradient

For deterministic policies $a = \mu_\theta(s)$ [4],

$$\nabla_\theta J = E_{s\sim d_\mu}\left[\nabla_\theta \mu_\theta(s)\, \nabla_a Q_\phi(s,a)\right]_{a=\mu_\theta(s)} \tag{4}$$

with TD target $y = r + \gamma\, Q_{\bar\phi}\left(s', \mu_{\bar\theta}(s')\right)$. Target networks use Polyak averaging.

### D. Twin Delayed DDPG (TD3)

TD3 [5] reduces overestimation via twin critics and target smoothing:

$$y = r + \gamma \min_{i\in\{1,2\}} \bar{Q}_{\bar\phi_i}(s', \mu_{\bar\theta}(s') + \epsilon) \tag{5}$$

$$\epsilon \sim \text{clip}(\mathcal{N}(0,\sigma^2), -c, c) \tag{6}$$

and *delays* actor/target updates.

### E. Proximal Policy Optimization (PPO)

PPO [6] maximizes a clipped surrogate objective for a stochastic actor:

$$\mathcal{L}_{PPO}(\theta) = E_t\left[min\left(r_t(\theta)\hat{A}_t, clip(r_t(\theta), 1-\varepsilon, 1+\varepsilon)\hat{A}_t\right)\right] \tag{7}$$

$$r_t(\theta) = \frac{\pi_\theta(a_t|s_t)}{\pi_{\theta_{\text{old}}}(a_t|s_t)} \tag{8}$$

where $\widehat{A_t}$ is an advantage estimate.

### F. SAC (Maximum Entropy)

SAC [7] augments the return with an entropy bonus to encourage exploration. The maximum-entropy objective and soft value are

$$J(\pi) = E_{(s,a)\sim\rho_\pi}\left[r(s,a) + \alpha\mathcal{H}\left(\pi(\cdot\,|s)\right)\right] \tag{9}$$

$$V^\pi(s) = E_{a\sim\pi}[Q^\pi(s,a) - \alpha log\pi(a|s)] \tag{10}$$

leading to the soft Bellman target

$$y = r + \gamma E_{a'\sim\pi(\cdot|s')}\left[min_i \bar{Q}_{\bar\phi_i}(s',a') - \alpha log\pi(a'|s')\right] \tag{11}$$

and a Tanh-squashed Gaussian policy via reparameterization; $\alpha$ can be tuned to a target entropy.

## III. RELATED WORK

**RL for aerial robots.** Early surveys established the algorithmic foundations of RL and DRL [1], [2],

[18]–[20], and recent reviews focused specifically on drones [21]. Within UAVs, DRL has been used broadly for high-level tasks such as navigation, path planning, and target tracking in complex or large-scale environments [22]–[27]. These works act on waypoints or velocity set-points rather than the vehicle's low-level actuation, and typically rely on simplified dynamics provided by gym-like simulators.

**Low-level attitude/thrust control.** Closer to our setting, DRL has been applied to attitude stabilization and thrust control [8]–[11], [28] typically commanding body moments and total thrust with instantaneous actuation and neglecting allocation and lags. Examples include cascaded designs [8], disturbance robustness [9], sim-to-real demonstrations [11], and on- vs. off-policy comparisons [10].

**End-to-end and task-driven control.** Aggressive and time-optimal flight has been shown with perception-to-control pipelines [12], and task-driven lines such as swarms and multi-tasking highlight agility [29], [30], but they are not primarily concerned with actuator/propulsor fidelity.

**Architectural priors and multi-agent control.** Inductive biases (symmetry/equivariance) improve sample efficiency and generalization; see equivariant policies/critics and multi-agent extensions for quadrotors [13], [14]. These are complementary to physical fidelity.

**Reproducibility and infrastructure.** The community has underscored the sensitivity of DRL results to implementation details and random seeds [15], spurring reliable libraries [16] and analyses of parallel/distributed training [31]. We follow these best practices in Sec. V.

**Positioning.** Unlike body-level abstractions [8]–[10], we expose policies to a realistic actuation pathway: body-wrench actions $\{T, \tau_x, \tau_y, \tau_z\}$ are mapped to per-rotor speeds via allocation that includes thrust/torque coupling, and actuator dynamics are retained. The approach remains algorithm-agnostic (SAC/TD3/DDPG/PPO) [4]–[7], providing a physics-aware testbed closer to hardware constraints.

## IV. SYSTEM MODEL & NOTATION

We model a rigid quadcopter with an inertial frame $\{i\}$ and a body-fixed frame $\{b\}$. Roll, pitch, and yaw follow the Z-Y-X (yaw-pitch-roll) convention with Euler angles $(\phi, \theta, \psi)$.

### A. Frames and ZYX Rotation

For brevity, let $c_\star = \cos(\star)$ and $s_\star = \sin(\star)$. The rotation from body to inertial is

$$\boldsymbol{R}_{ib} = \begin{bmatrix} c_\psi c_\theta & c_\psi s_\theta s_\phi - s_\psi c_\phi & c_\psi s_\theta c_\phi + s_\psi s_\phi \\ s_\psi c_\theta & s_\psi s_\theta s_\phi + c_\psi c_\phi & s_\psi s_\theta c_\phi - c_\psi s_\phi \\ -s_\theta & c_\theta s_\phi & c_\theta c_\phi \end{bmatrix}$$

$$\boldsymbol{R}_{bi} = \boldsymbol{R}_{ib}^\top \tag{12}$$

Gravity is specified in $\{i\}$ and projected to $\{b\}$ as

$$\boldsymbol{g}_i = \begin{bmatrix} 0 \\ 0 \\ -g \end{bmatrix}, \quad \boldsymbol{g}_b = \boldsymbol{R}_{bi}\boldsymbol{g}_i \tag{13}$$

### B. States, Inputs, Outputs

The 12-dimensional state, low-level input, and full-state output are

$$\boldsymbol{x} = \begin{bmatrix} P \\ Q \\ R \\ \phi \\ \theta \\ \psi \\ U \\ V \\ W \\ X \\ Y \\ Z \end{bmatrix}, \quad \boldsymbol{u} = \begin{bmatrix} T \\ \tau_x \\ \tau_y \\ \tau_z \end{bmatrix}, \quad \boldsymbol{y}(t) = \boldsymbol{x}(t) \tag{14}$$

### C. Kinematics: Euler and Angular Rates

Let $\boldsymbol{\omega}_b = [P \;\; Q \;\; R]^\top$. The mapping between Euler-angle rates and body angular rates is

$$\begin{bmatrix} \dot{\phi} \\ \dot{\theta} \\ \dot{\psi} \end{bmatrix} = \boldsymbol{H}(\phi, \theta)\boldsymbol{\omega}_b$$

$$\boldsymbol{H}(\phi, \theta) = \begin{bmatrix} 1 & sin\phi tan\theta & cos\phi tan\theta \\ 0 & cos\phi & -sin\phi \\ 0 & \frac{sin\phi}{cos\theta} & \frac{cos\phi}{cos\theta} \end{bmatrix} \tag{15}$$

### D. Forces and Moments (motor/thrust level)

Thrust acts along body $+z_b$:

$$\boldsymbol{F}_b = \begin{bmatrix} 0 \\ 0 \\ \sum_{i=1}^{4} c_{t,i}\, w_i^2 \end{bmatrix} \tag{16}$$

Optional external disturbances given in $\{i\}$ are projected to $\{b\}$:

$$\boldsymbol{F}_{dist}^{b} = \boldsymbol{R}_{bi} \begin{bmatrix} F_x \\ F_y \\ F_z \end{bmatrix} \tag{17}$$

Rotor gyroscopic torques (in code form, $w_i$ in RPM; $J_m$ is motor inertia) are

$$\boldsymbol{\tau}_{gyro} = \begin{bmatrix} Q\,J_m \frac{2\pi}{60}(-w_1 - w_3 + w_2 + w_4) \\ P\,J_m \frac{2\pi}{60}(w_1 + w_3 - w_2 - w_4) \\ 0 \end{bmatrix} \quad (18)$$

Total body moments:

$$\boldsymbol{M}_b = \boldsymbol{D}\begin{bmatrix} w_1^2 \\ w_2^2 \\ w_3^2 \\ w_4^2 \end{bmatrix} + \boldsymbol{\tau}_{gyro} + \begin{bmatrix} \tau_x \\ \tau_y \\ \tau_z \end{bmatrix} \quad (19)$$

where $D \in R^{3\times 4}$ collects the moment/drag coefficients (often denoted by the "$dctcq$" matrix).

## E. Rigid-Body Dynamics

*a) Angular dynamics:*

With inertia $\boldsymbol{J}_b = diag(J_x, J_y, J_z)$ and

$$[\boldsymbol{\omega}_b]_\times = \begin{bmatrix} 0 & -R & Q \\ R & 0 & -P \\ -Q & P & 0 \end{bmatrix} \quad (20)$$

the rotational dynamics are

$$\dot{\boldsymbol{\omega}}_b = \boldsymbol{J}_b^{-1}(\boldsymbol{M}_b - [\boldsymbol{\omega}_b]_\times \boldsymbol{J}_b \boldsymbol{\omega}_b) \quad (21)$$

*b) Translational dynamics:*

Let $\boldsymbol{v}_b = [\,U\ \ V\ \ W]^\top$ and $m$ the mass:

$$\dot{\boldsymbol{v}}_b = \frac{1}{m}\left(\boldsymbol{F}_b + \boldsymbol{F}_{dist}^b\right) + \boldsymbol{g}_b - [\boldsymbol{\omega}_b]_\times \boldsymbol{v}_b \quad (22)$$

*c) Position kinematics:*

$$\dot{\boldsymbol{p}}_i = \boldsymbol{R}_{ib}\boldsymbol{v}_b, \quad \boldsymbol{p}_i = [X\ \ Y\ \ Z]^\top \quad (23)$$

*d) State derivative (12 states):*

$$\dot{\boldsymbol{x}} = \begin{bmatrix} \dot{\boldsymbol{\omega}}_b \\ \boldsymbol{H}(\phi,\theta)\boldsymbol{\omega}_b \\ \dot{\boldsymbol{v}}_b \\ \boldsymbol{R}_{ib}\boldsymbol{v}_b \end{bmatrix} \quad (24)$$

## F. Action-to-RPM Allocation

The RL action $a = [a_1\ \ a_2\ \ a_3\ \ a_4]^\top \in [-1,1]^4$ is mapped to total thrust and body torques:

$$T = T_{hover} + \alpha_T T_{hover} a_1, \quad T_{hover} = mg \quad (25)$$

$$\begin{bmatrix} \tau_x \\ \tau_y \\ \tau_z \end{bmatrix} = \begin{bmatrix} \tau_{scale,x} & 0 & 0 \\ 0 & \tau_{scale,y} & 0 \\ 0 & 0 & \tau_{scale,z} \end{bmatrix} \begin{bmatrix} a_2 \\ a_3 \\ a_4 \end{bmatrix} \quad (26)$$

Let

$$\boldsymbol{u}_{alloc} = \begin{bmatrix} T & \tau_x & \tau_y & \tau_z \end{bmatrix}^\top$$

$$\boldsymbol{M} = \begin{bmatrix} c_{t,1} & c_{t,2} & c_{t,3} & c_{t,4} \\ dctcq_{1,1} & dctcq_{1,2} & dctcq_{1,3} & dctcq_{1,4} \\ dctcq_{2,1} & dctcq_{2,2} & dctcq_{2,3} & dctcq_{2,4} \\ dctcq_{3,1} & dctcq_{3,2} & dctcq_{3,3} & dctcq_{3,4} \end{bmatrix} \quad (27)$$

We solve the allocation in $w_i^2$ ($\mathrm{RPM}^2$) by a least-squares map

$$\begin{bmatrix} w_1^2 \\ w_2^2 \\ w_3^2 \\ w_4^2 \end{bmatrix} = \boldsymbol{M}^+ \boldsymbol{u}_{alloc} \quad (28)$$

$$w_i = \sqrt{max\left(w_i^2, 0\right)}, \quad w_i \in [w_{min}, w_{max}] \quad (29)$$

where $M^+$ is the Moore–Penrose pseudo-inverse (used when $M$ is not square or poorly conditioned). The clipping to $[w_{min}, w_{max}]$ enforces actuator limits.

## G. Motor First-Order Dynamics

Each commanded RPM passes a first-order lag before acting on forces/torques:

$$T_m\,\dot{y}(t) + y(t) = u(t), \quad \frac{Y(s)}{U(s)} = \frac{1}{T_m s + 1} \quad (30)$$

with time constant $T_m = 0.076\ \mathrm{s}$ in our configuration. Here $u(t)$ is the commanded RPM and $y(t)$ the realized RPM used in the thrust/torque computations above.

# V. METHODOLOGY

We describe how the DRL agent interacts with the quadcopter plant defined in Sec. IV, emphasizing observation/action design, reward shaping, and the training pipeline.

## A. Observation Space

The agent receives a structured observation vector comprising:

- Position $\mathbf{p_t} = [X_t\ \ Y_t\ \ Z_t]^\top$ and error ($\mathbf{e_t} = \mathbf{p_t} - \mathbf{p}^*$);
- Euler angles ($\boldsymbol{\eta_t} = [\phi_t\ \ \theta_t\ \ \psi_t]^\top$), body-frame linear velocity ($\mathbf{v_t^b} = [U_t\ \ V_t\ \ W_t]^\top$) and body rates ($\boldsymbol{\omega_t} = [P_t\ \ Q_t\ \ R_t]^\top$)
- Previous action ($\mathbf{a_{t-1}}$) (provides control history for rotor inertia/lag).

## B. Action Space

Actions are continuous and directly correspond to thrust and body torques:

$$\boldsymbol{a} = \left[a_T\ \ a_{\tau_x}\ \ a_{\tau_y}\ \ a_{\tau_z}\right]$$

which are mapped to per-rotor speeds through the allocation model in Sec. IV. This ensures physical feasibility and respects actuator limits.

## C. Reward Design

To drive the agent toward the goal while keeping the vehicle stable, we shape the reward with (i) a distance-based positive term, (ii) attitude and rate penalties for stability, and (iii) a small survival bonus.

Let $\boldsymbol{p}_t = [x_t\ y_t\ z_t]^\top$, $\boldsymbol{p}^* = [x_{\text{ref}}\ y_{\text{ref}}\ z_{\text{ref}}]^\top$, $d_t = \|\boldsymbol{p}_t - \boldsymbol{p}^*\|_2$, body-frame linear velocity $\boldsymbol{v}_t^b = [U_t\ V_t\ W_t]^\top$, body rates $(P_t, Q_t, R_t)$, and Euler angles $(\phi_t, \theta_t, \psi_t)$ (rad). Define the distance shaping term:

$$r_d(d_t) = \begin{cases} w_d e^{-\beta d_t}, & d_t \le d_{th} \\ -2d_t, & d_t > d_{th} \end{cases} \quad (31)$$

and the total reward:

$$r_t = r_d(d_t) - \kappa_\phi|\phi_t| - \kappa_\theta|\theta_t| - \kappa_\psi|\psi_t| \\ -\lambda_v\|v_t^b\|_2^2 - \lambda_\omega(P_t^2 + Q_t^2 + \rho^2 R_t^2) + b_{live} \quad (32)$$

**Weights used in experiments.** We set $w_d = 30$, $\beta = 3$, $d_{\text{th}} = 0.5\,\text{m}$, $\lambda_v = 0.01$, $\lambda_\omega = 0.003$, $\rho = 0.2$, and $b_{live} = 1$. The attitude penalties share the same gain:

$$\kappa_\phi = \kappa_\theta = \kappa_\psi = 2\frac{180}{\pi}$$

## D. Training Pipeline

We adopt an end-to-end DRL training framework implemented in MATLAB/Simulink:

1) The agent interacts with the S-function-based plant at each timestep.
2) Observations are computed, actions are generated, and mapped to motor dynamics.
3) Rewards are collected and used for policy updates.

We evaluate four algorithms—DDPG, TD3, PPO, and SAC—under identical network structures and hyperparameters for fair comparison.

# VI. EXPERIMENTS

This section documents the experimental setup used to evaluate the proposed approach. We detail the Simulink-in-the-loop training architecture, plant configuration, observation/action spaces, reward shaping, algorithms and hyperparameters, curriculum, hardware, and evaluation protocol. All dynamics and notation follow Sec. IV.

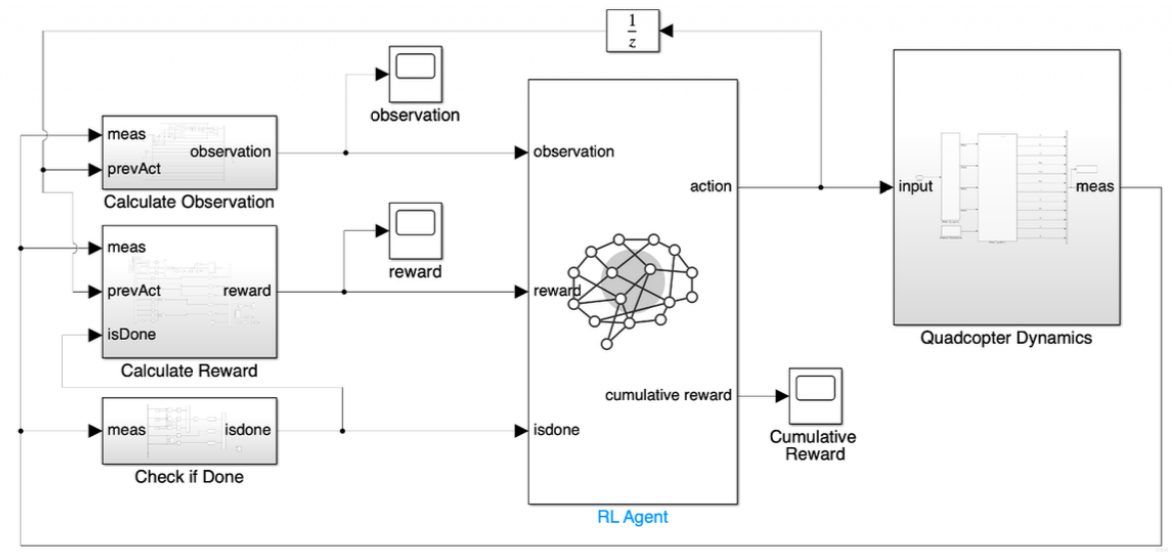

Fig. 1. Simulink-in-the-loop training architecture. The plant block *Quadcopter Dynamics* is a 12-state Level-2 S-Function. At each step, *Calculate Observation* assembles the observation vector (including the previous action via the unit-delay), the *RL Agent* outputs low-level controls (total thrust and body torques), and the action is applied to the plant. *Calculate Reward* computes the scalar reward and *Check if Done* determines episode termination. The measurement bus closes the loop and the scope accumulates the episode return.

TABLE I
CORE SIMULATION PARAMETERS

| Parameter | Value | Unit / Note |
|---|---|---|
| $g$ | 9.810 | m/s$^2$ |
| $m$ | 1.0230 | kg |
| $d$ | 0.2223 | m (arm length) |
| $J_x, J_y, J_z$ | 0.0095, 0.0095, 0.0186 | kg·m$^2$ |
| $J_m$ | $3.7882 \times 10^{-6}$ | kg·m$^2$ (rotor) |
| $T_m$ | 0.0760 | s (motor time constant) |
| $c_t$ | $1.4865 \times 10^{-7}$ | N/RPM$^2$ (per rotor) |
| $c_q$ | $2.9250 \times 10^{-9}$ | N·m/RPM$^2$ |
| $w_{\min}, w_{\max}$ | 1500, 9000 | RPM |
| $\alpha_T$ | 0.20 | thrust range (rel. hover) |
| $\tau_{\text{scale}}$ | [0, 0.001, 0] | N·m limits (x,y,z) |

## A. Training Architecture

Fig. 1 shows the closed-loop data flow per simulation step ($T_s = 0.005s$, horizon $T_f = 10s$).

## B. Plant & Model Configuration

We use a rigid-body quadcopter with ZYX rotation, thrust along body $+z_b$, gravity $g_i = [0,0,-g]^\top$, and actuator lag $T_m = 0.076s$. Allocation solves for squared rotor speeds given total thrust and body torques (Sec. IV). Table I lists key parameters.

## C. Observation & Action Spaces

Observations: position error to target, Euler angles $(\phi, \theta, \psi)$, body velocities $[U, V, W]$, body rates $[P, Q, R]$, and previous action $\mathbf{a_{t-1}}$.

Actions: $\boldsymbol{a} = \left[a_T\ a_{\tau_x}\ a_{\tau_y}\ a_{\tau_z}\right]$, mapped to $\left[T\ \tau_x\ \tau_y\ \tau_z\right]$ and then to per-rotor RPM (least-squares in $\omega_i^2$ with clamping).

## D. Reward & Termination

We use a shaped reward (Eq. (32)) combining exponential distance shaping, angle/velocity penalties, and a small living bonus; termination triggers on out-of-bound states or reaching a goal ball with dwell time.

## E. Algorithms & Hyperparameters

We compare SAC, TD3, DDPG (and PPO in thrust-only Stage 1) under identical network sizes and optimizer settings; buffers of $10^6$, mini-batch 256, discount $\gamma = 0.99$, target update $\tau = 0.005$, learning rate $3 \times 10^{-4}$. SAC uses twin critics and an entropy target $-\dim(\mathcal{A})$; TD3 uses twin critics and target policy smoothing.

## F. Curriculum & Initialization

- **Stage 1**: thrust-only hover ($\tau_x = \tau_y = \tau_z = 0$) for 5k episodes with randomized initial $X, Z$.
- **Stage 2**: unlocks one moment channel (e.g., $\tau_y$) and trains 18k episodes to reach $(1, 0, 1)$m, with

$$X \sim \mathcal{N}(0,1^2),\ Y = 0,\ Z \sim \mathcal{N}(0,1^2).$$

## G. Hardware & Parallelization

MATLAB Parallel Computing Toolbox with up to 8 workers; GPU when available. We limit CPU saturation to reduce OS-level nondeterminism.

# VII. RESULTS

Unless otherwise stated, all experiments follow the common configuration of Sec. VI (sample time $T_s = 5\,\text{ms}$, horizon $T_f = 10\,\text{s}$; identical observations/actions and reward per (31)–(32)). Every 100 episodes, an evaluator executes five noise-free rollouts; markers in the plots denote these evaluation returns.

## A. Stage 1: Thrust-only hover

Fig. 2 indicates consistent improvement for the off-policy baselines. **TD3** exhibits a near-monotonic increase in training return after $\approx 10^3$ episodes and attains the highest final reward. **SAC** follows closely, with slightly slower early growth yet comparable terminal performance. **DDPG** improves rapidly at first but displays pronounced oscillations and a lower asymptote. **PPO** remains near zero throughout and shows no persistent upward trend within the given budget. Evaluation markers mirror the ordering of training curves, suggesting that gains are not artifacts of exploration noise or replay dynamics.

## B. Stage 2: Thrust plus single-axis torque

With a translated target and the pitch-torque ($\tau_y$) channel enabled, returns start negative and increase more gradually (Fig. 3). **TD3** maintains the best sample efficiency, producing a steady ascent over the entire training window. **SAC** is more conservative in the initial phase—consistent with entropy-regularized exploration—but accelerates in the latter half and reaches performance comparable to, and occasionally exceeding on evaluation rollouts, the **TD3** baseline. **DDPG** remains noisy and does not surpass the twin-critic methods under this setting. (**PPO** is excluded in Stage 2 due to its poor convergence in Stage 1.)

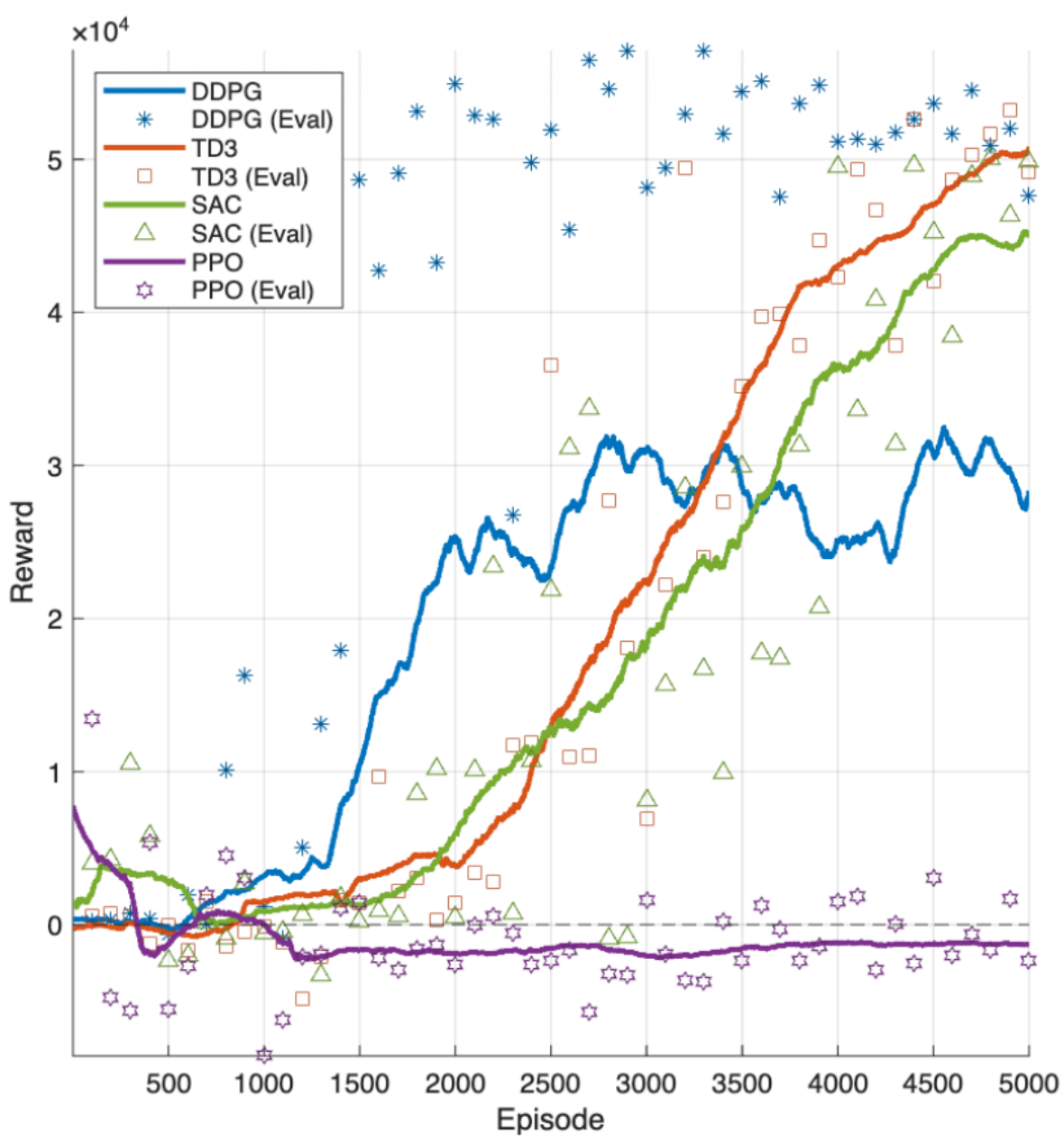

Fig. 2. **Stage-1** Training Curve Comparison. Lines: training; markers: eval.

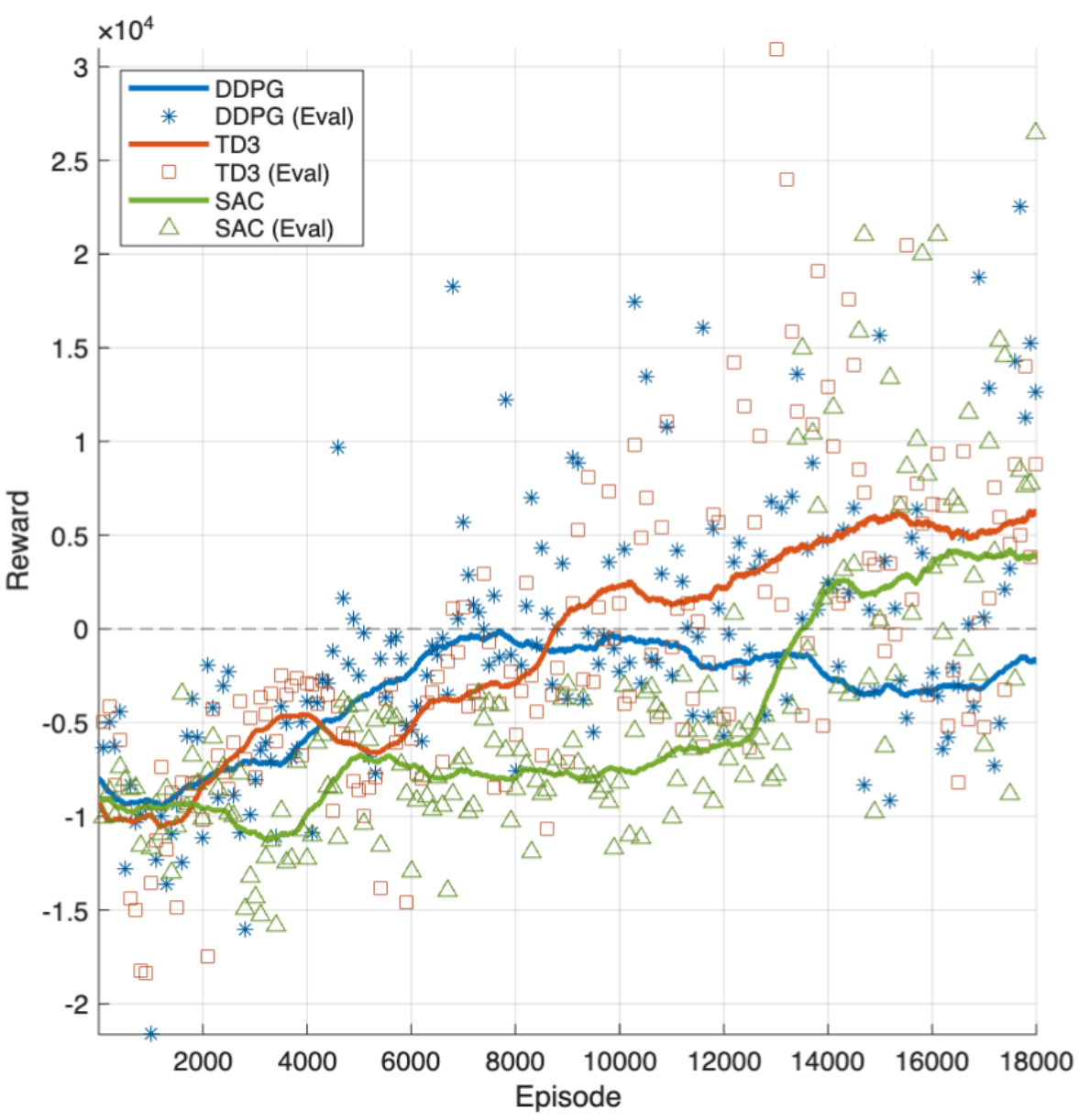

Fig. 3. **Stage-2** Training Curve Comparison. Lines: training; markers: eval.

## C. Trajectory-level assessment

The planar projections and 3D trajectory of the best **SAC** agent in Stage 2 (Fig. 5) demonstrate a direct approach toward the goal with negligible lateral excursion: $Y$ remains essentially constant, while $X$ advances toward $1\,\text{m}$ and $Z$ ascends to $1\,\text{m}$. Time histories (Fig. 4) show a bounded overshoot in $Z$ followed by damped oscillations and settling near the reference; $X$ exhibits a smooth rise with a mild corrective phase near arrival; $Y$ drift is numerically negligible (note the axis scale of $10^{-8}$). Table II quantitatively supports these observations.

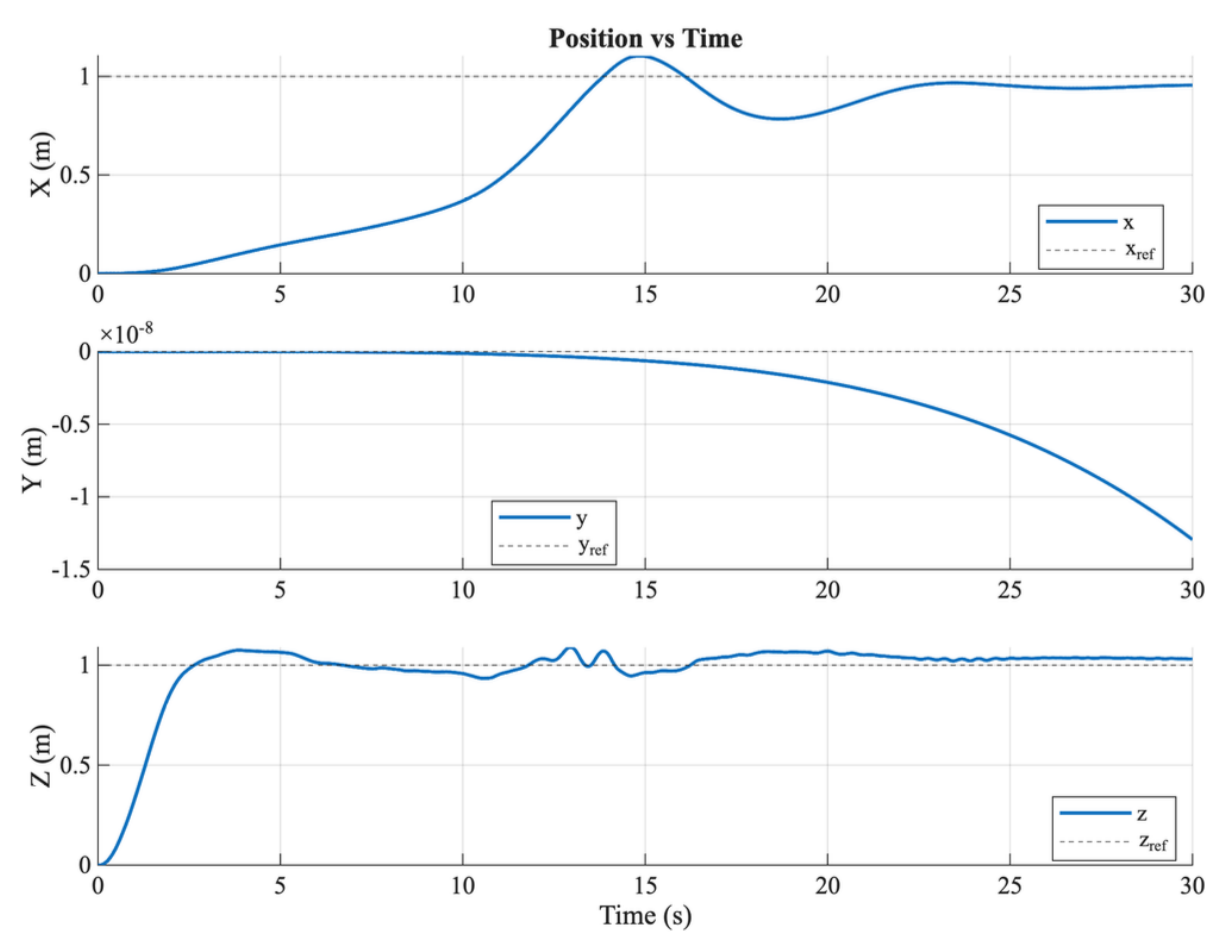

Fig. 4. Position vs time of the best-performing SAC agent (Stage 2).

## VIII. CONCLUSION

This work advanced low-level quadcopter control by coupling continuous body-wrench actions with an explicit thrust/torque allocation and first-order rotor/motor dynamics in a Simulink-in-the-loop setting. Empirically, this added physical fidelity makes the learning problem harder—not easier—by introducing actuator lags and allocation nonlinearities that increase variance and slow improvement relative to body-level abstractions. Within a two-stage curriculum (thrust-only hover; thrust plus a single torque axis), off-policy methods converged most reliably, with TD3 showing the fastest progress and SAC achieving comparable final performance with stronger evaluation consistency; DDPG and PPO underperformed under the same budget. However, training remains brittle and we were unable to obtain a stable Stage-3 policy for fully coupled 3D attitude and position. Future work will remain fully end-to-end and investigate multi-agent coordination, richer observation and reward formulations, and architecture-level advances to reduce sensitivity and enable reliable full 3D control under coupled actuator dynamics.

## ACKNOWLEDGMENT

This project was supported by the National Science and Technology Council of Taiwan (112-2222-E-006-001-MY3).

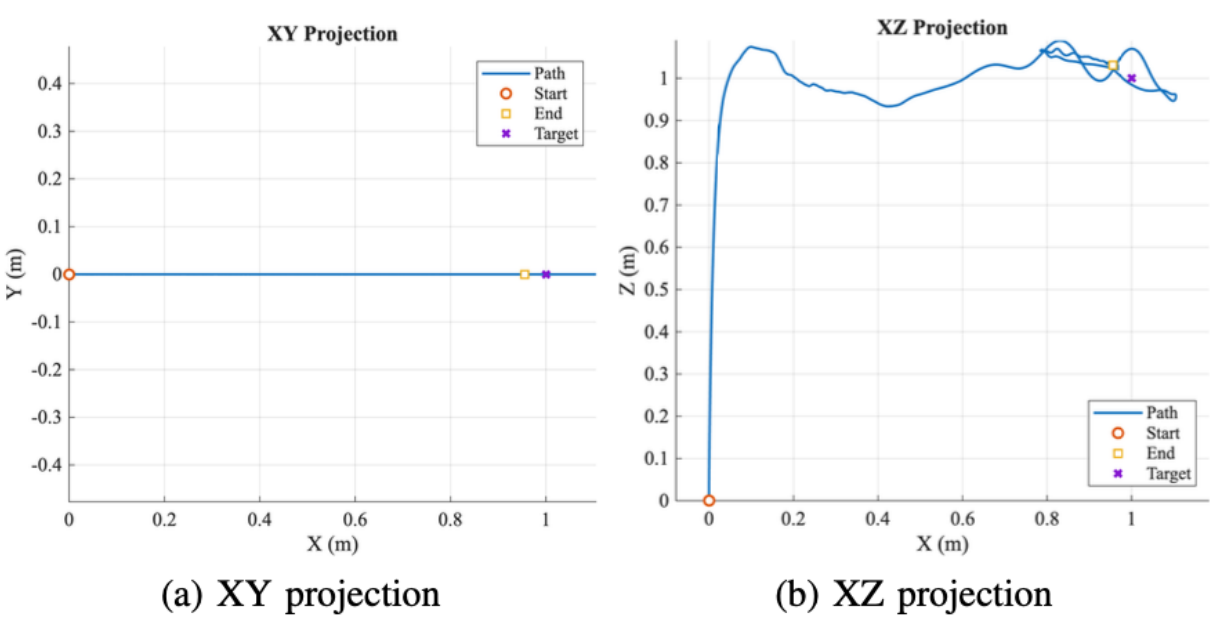

(a) XY projection (b) XZ projection

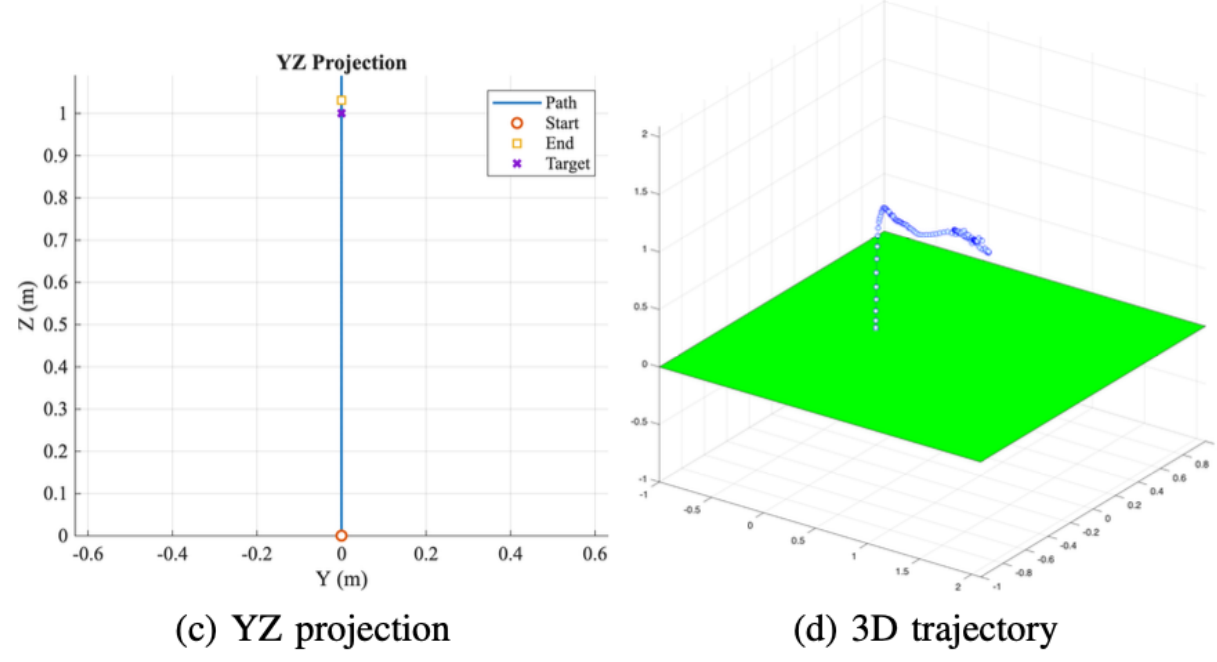

(c) YZ projection (d) 3D trajectory

Fig. 5. Planar and 3D trajectories of the best-performing SAC agent (Stage 2).

TABLE II
QUANTITATIVE EVALUATION METRICS OF THE BEST-PERFORMING SAC AGENT (STAGE 2)

| Metric | X | Y | Z | 3D |
|---|---|---|---|---|
| RMSE (m) | 0.526 | 0.000 | 0.199 | 0.562 |
| MAE (m) | 0.379 | 0.000 | 0.084 | 0.402 |
| **Steady-state RMS (1 s)** | 0.0560 m | | | |
| **Steady-state mean (1 s)** | 0.0560 m | | | |
| First enter 0.5 m band | 11.165 s | | | |
| Settle time (stay within 0.05 m) | 11.165 s | | | |
| Reach goal (0.10 m) | 13.305 s | | | |
| Path length | 3.058 m | | | |
| Mean speed | 0.102 m/s | | | |
| Peak speed | 0.606 m/s | | | |
| Final position $(x, y, z)$ | (0.956, 0.000, 1.031) m | | | |
| Final position error | 0.054 m | | | |

## REFERENCES

[1] K. Arulkumaran, M. P. Deisenroth, M. Brundage, and A. A. Bharath, "Deep reinforcement learning: A brief survey," IEEE signal processing magazine, vol. 34, no. 6, pp. 26–38, 2017.
[2] R. S. Sutton, A. G. Barto et al., Reinforcement learning: An introduction. MIT press Cambridge, 1998, vol. 1, no. 1.
[3] V. Mnih, K. Kavukcuoglu, D. Silver, A. Graves, I. Antonoglou, D. Wierstra, and M. Riedmiller, "Playing atari with deep reinforcement learning," arXiv preprint arXiv:1312.5602, 2013.
[4] T. P. Lillicrap, J. J. Hunt, A. Pritzel, N. Heess, T. Erez, Y. Tassa, D. Silver, and D. Wierstra, "Continuous control with deep reinforcement learning," arXiv preprint arXiv:1509.02971, 2015.
[5] S. Fujimoto, H. Hoof, and D. Meger, "Addressing function approxi- mation error in actor-critic methods," in International conference on machine learning. PMLR, 2018, pp. 1587–1596.
[6] J. Schulman, F. Wolski, P. Dhariwal, A. Radford, and O. Klimov, "Prox- imal policy optimization algorithms," arXiv preprint arXiv:1707.06347, 2017.
[7] T. Haarnoja, A. Zhou, P. Abbeel, and S. Levine, "Soft actor-critic: Off- policy maximum entropy deep reinforcement learning with a stochastic actor," in International conference on machine learning. Pmlr, 2018, pp. 1861–1870.
[8] W. Koch, R. Mancuso, R. West, and A. Bestavros, "Reinforcement learning for uav attitude control," ACM Transactions on Cyber-Physical Systems, vol. 3, no. 2, pp. 1–21, 2019.
[9] A. M. Deshpande, A. A. Minai, and M. Kumar, "Robust deep rein- forcement learning for

quadcopter control," IFAC-PapersOnLine, vol. 54, no. 20, pp. 90–95, 2021.
[10] Z. Jiang and A. F. Lynch, "Quadrotor motion control using deep reinforcement learning," Journal of Unmanned Vehicle Systems, vol. 9, no. 4, pp. 234–251, 2021.
[11] T.-D. Do, N. X. Mung, and S. K. Hong, "Deep reinforcement learning- based quadcopter controller: A practical approach and experiments," arXiv preprint arXiv:2406.08815, 2024.
[12] R. Ferede, C. De Wagter, D. Izzo, and G. C. De Croon, "End-to-end reinforcement learning for time-optimal quadcopter flight," in 2024 IEEE International Conference on Robotics and Automation (ICRA). IEEE, 2024, pp. 6172–6177.
[13] B. Yu and T. Lee, "Equivariant reinforcement learning for quadrotor uav," in 2023 American Control Conference (ACC). IEEE, 2023, pp. 2842–2847.
[14] ——, "Multi-agent reinforcement learning for the low-level control of a quadrotor uav," in 2024 American Control Conference (ACC). IEEE, 2024, pp. 1537–1542.
[15] P. Henderson, R. Islam, P. Bachman, J. Pineau, D. Precup, and D. Meger, "Deep reinforcement learning that matters," in Proceedings of the AAAI conference on artificial intelligence, vol. 32, no. 1, 2018.
[16] A. Raffin, A. Hill, A. Gleave, A. Kanervisto, M. Ernestus, and N. Dor- mann, "Stable-baselines3: Reliable reinforcement learning implementa- tions," Journal of machine learning research, vol. 22, no. 268, pp. 1–8, 2021.
[17] P. Ladosz, L. Weng, M. Kim, and H. Oh, "Exploration in deep rein- forcement learning: A survey," Information Fusion, vol. 85, pp. 1–22, 2022.
[18] L. P. Kaelbling, M. L. Littman, and A. W. Moore, "Reinforcement learning: A survey," Journal of artificial intelligence research, vol. 4, pp. 237–285, 1996.
[19] V. Franc¸ois-Lavet, P. Henderson, R. Islam, M. G. Bellemare, J. Pineau et al., "An introduction to deep reinforcement learning," Foundations and Trends® in Machine Learning, vol. 11, no. 3-4, pp. 219–354, 2018.
[20] X. Wang, S. Wang, X. Liang, D. Zhao, J. Huang, X. Xu, B. Dai, and Q. Miao, "Deep reinforcement learning: A survey," IEEE Transactions on Neural Networks and Learning Systems, vol. 35, no. 4, pp. 5064– 5078, 2022.
[21] A. T. Azar, A. Koubaa, N. Ali Mohamed, H. A. Ibrahim, Z. F. Ibrahim, M. Kazim, A. Ammar, B. Benjdira, A. M. Khamis, I. A. Hameed et al., "Drone deep reinforcement learning: A review," Electronics, vol. 10, no. 9, p. 999, 2021.
[22] C. Wang, J. Wang, Y. Shen, and X. Zhang, "Autonomous navigation of uavs in large-scale complex environments: A deep reinforcement learning approach," IEEE Transactions on Vehicular Technology, vol. 68, no. 3, pp. 2124–2136, 2019.
[23] C. Wang, J. Wang, J. Wang, and X. Zhang, "Deep-reinforcement- learning-based autonomous uav navigation with sparse rewards," IEEE Internet of Things Journal, vol. 7, no. 7, pp. 6180–6190, 2020.
[24] C. Yan, X. Xiang, and C. Wang, "Towards real-time path planning through deep reinforcement learning for a uav in dynamic environ- ments," Journal of Intelligent & Robotic Systems, vol. 98, no. 2, pp. 297–309, 2020.
[25] A. Rodriguez-Ramos, C. Sampedro, H. Bavle, P. De La Puente, and P. Campoy, "A deep reinforcement learning strategy for uav autonomous landing on a moving platform," Journal of Intelligent & Robotic Systems, vol. 93, no. 1, pp. 351–366, 2019.
[26] K. K. Nguyen, T. Q. Duong, T. Do-Duy, H. Claussen, and L. Hanzo, "3d uav trajectory and data collection optimisation via deep reinforcement learning," IEEE Transactions on Communications, vol. 70, no. 4, pp. 2358–2371, 2022.
[27] B. Ma, Z. Liu, W. Zhao, J. Yuan, H. Long, X. Wang, and Z. Yuan, "Target tracking control of uav through deep reinforcement learning," IEEE Transactions on Intelligent Transportation Systems, vol. 24, no. 6, pp. 5983–6000, 2023.
[28] J. Hwangbo, I. Sa, R. Siegwart, and M. Hutter, "Control of a quadrotor with reinforcement learning," IEEE Robotics and Automation Letters, vol. 2, no. 4, pp. 2096–2103, 2017.
[29] S. Batra, Z. Huang, A. Petrenko, T. Kumar, A. Molchanov, and G. S. Sukhatme, "Decentralized control of quadrotor swarms with end-to-end deep reinforcement learning," in Conference on robot learning. PMLR, 2022, pp. 576–586.
[30] J. Xing, I. Geles, Y. Song, E. Aljalbout, and D. Scaramuzza, "Multi-task reinforcement learning for quadrotors," IEEE Robotics and Automation Letters, 2024.
[31] Z. Liu, X. Xu, P. Qiao, and D. Li, "Acceleration for deep reinforcement learning using parallel and distributed computing: A survey," ACM Computing Surveys, vol. 57, no. 4, pp. 1–35, 2024.